\documentclass[11pt,a4paper]{article}
\usepackage[hyperref]{naaclhlt2018}
\usepackage{enumitem}
\usepackage{times}
\usepackage{latexsym}
\usepackage{amsmath}
\usepackage{url}

\aclfinalcopy 

\newcommand\blfootnote[1]{%
  \begingroup
  \renewcommand\thefootnote{}\footnote{#1}%
  \addtocounter{footnote}{-1}%
  \endgroup
}

\title{Selecting Machine-Translated Data for Quick Bootstrapping of a Natural Language Understanding System}

\author{Judith Gaspers \\
  Amazon\\
  Aachen, Germany\\
  {\tt gaspers@amazon.de} \\\And
Penny Karanasou \\
  Amazon \\
  Cambridge, UK \\
  {\tt pkarana@amazon.co.uk }\\\And
Rajen Chatterjee\\\
  University of Trento \\
  Trento, Italy\\
  {\tt chatterjee@unitn.it} \\}
\date{}

\begin{document}
\maketitle
\begin{abstract}
This paper investigates the use of Machine Translation (MT) to bootstrap a Natural Language Understanding (NLU) system for a new language for the use case of a large-scale voice-controlled device. The goal is to decrease the cost and time needed to get an annotated corpus for the new language, while still having a large enough coverage of user requests. Different methods of filtering MT data in order to keep utterances that improve NLU performance and language-specific post-processing methods are investigated.
These methods are tested in a large-scale NLU task with translating around 10 millions training utterances from English to German. The results show a large improvement for using MT data over a grammar-based and over an in-house data collection baseline, while reducing the manual effort greatly. Both filtering and post-processing approaches improve results further. 
\blfootnote{The author Rajen Chatterjee conducted the work for this paper during an internship at Amazon, Cambridge, UK}
 \end{abstract}


\section{Introduction}

In recent years, there has been growing interest in voice-controlled devices, such as Amazon Alexa or Google home. This success makes the quick bootstrapping of corresponding systems, including NLU models, for new languages a prioritised goal. However, building a new NLU model for each language from scratch and gathering the necessary annotated corpora implies a significant amount of human time and effort both by annotators and scientists. In addition, this procedure is not scalable to supporting an increasing number of languages. On the other hand, a large amount of data is usually available for the language(s) that are already supported. Leveraging this source of data seems an obvious solution. In this paper, we investigate the use of Machine Translation to translate existing data sources to a new target language and use them to bootstrap an NLU system for this target language. 

A common procedure for data gathering for a new language starts by some grammar-generated data. Significant time and effort is consumed at this stage by language specialists to build grammars that offer a good coverage needed for a first working system. Once this first system reaches a certain performance threshold, it can be shared with beta users. This step allows more data that cover real user's queries to be generated. All existing data sources are then used to train the system that will be released to the final customers, once a new higher performance threshold is reached. Finally, when the system is released to the customers, customer data become available. Beta and customer data better cover the user utterances than grammar-generated data and are, thus, valuable for the development of a good and generalisable NLU system. However, it takes a significant amount of time and human annotation effort in order to have enough annotated beta, and later customer data, needed to build a good NLU system. Furthermore, having a system robust to new domains and features is very challenging and requires data with a wide coverage.

Machine Translation can be a useful tool for the quick expansion to new languages by automatically translating customer data from existing resources to new languages. This could decrease significantly the time needed to develop an NLU system that replies well to customer queries and is robust to new features. In this paper, we work with a large-scale system where around 10 millions annotated customer data are available for US English with a wide coverage of domains and features. We use this corpus to augment the training data of a new language. In particular, we will present our experiments on applying our technique to bootstrap a German NLU system based on existing US English training data.

In addition, we explore ways to choose the ``good'' translations from the translated ones, i.e. the ones that improve the NLU performance. The investigated methods fall in the following categories. First, we investigate filtering based on MT quality. This method makes use of scores generated by the MT model to assign the quality of translations. 
The second method explores improving the NLU performance by making sure the filtered translations keep the semantic information required by the NLU system. In this case, the matching of the NLU labels after a backward translation task is used as the filtering criterion. Lastly, some language-specific post-processing is applied on the translation output. This includes resampling data with catalogues of the new language. Another post-processing step applied is to keep the original (EN) version of certain slots that the users tend to leave untranslated.   

This paper is organised as follows. In Section \ref{sec:back}, we give an overview of related literature. In Section \ref{sec:methods}, we present the methods for MT filtering for bootstrapping a new language while improving NLU performance. Next, we detail the experimental setup in section \ref{sec:exp}, including details on the used NLU and MT systems as well as the monolingual and bilingual corpora used. Afterwards, we present results in Section \ref{sec:res} before concluding the paper in Section \ref{sec:concl} .


\section{Background work}
\label{sec:back}

Many efforts to avoid or minimize this manual work have been made in the last few years using transfer learning, active learning and semi-supervised training. One of the successful approaches has been making use of an MT system to obtain annotated corpora. The results of such works depend on the availability of an MT system (general-purpose or in-domain), on the quality of the acquired translations and on the precision of NLU label-word alignment when passing from one language to another. \citet{Garcia2012} combine multiple online general-purpose translation systems to achieve transferability between French and Spanish for a dialog system. \citet{Jabaian2011} study phrase-based translation as an alternative to Conditional Random Fields (CRF) to keep NLU label-word alignment info in the decoding process. \citet{Lefevre2010} propose the  Semantic Tuple Classifiers (STC) model without any need for alignment information. \citet{Servan2010} translate the conceptual segments (i.e. NLU labeled) separately to maintain the chunking between source and target language but at the cost of degrading the translation quality. 

There is a wide literature on assessing the MT quality. Evaluating the quality of MT output has been a topic in the Workshop of Machine Translation (WMT) since its beginnings \cite{WMT2006} and a separate task since 2008 (``Shared Evaluation Task'' \cite{WMT2008}). Since 2012 a more specific ``Quality Estimation Task'' \cite{WMT2012} appears with a focus on deciding whether a translation is good and how to filter out translations that are not good enough. In addition, in 2017 \cite{WMT2017} other related topics appear including post-editing and bandit learning as specific tasks of correcting errors and improving MT quality by learning from feedback. A straight-forward method is using human translated data as the true reference and correct MT errors using this ground truth. Automatic Post-Editing (APE) can also improve MT quality by modifying MT output to the correct version \cite{Chatterjee2017}. Bandit learning \cite{Sokolov2017} replaces human reference and post-edits by a weak user's feedback. This feedback is introduced in the training process in a reinforcement learning framework updating the gradient to maximize the rewards corresponding to user's feedback.

However, all previous method focus on improving MT quality (i.e. BLEU score) and not the NLU task of interest. \citet{Jabaian2011} add noise to translation data and use translation post-editing to increase the robustness of NLU to translation errors. Other methods include measuring the probability of a translated utterance by applying a target language LM, i.e. measuring if a translated utterance is typical, or computing the likelihood that an alignment between the source and the translated utterance is correct, as \citet{Klinger2015} explore for the sentiment analysis task. We will do something similar in this paper by using directly the MT scores (alignment, translation and language model scores) as a measure of MT quality independent of the NLU tasks. In addition, we explore and extend a different approach for filtering which was presented by \citet{Misu2012}. In order to select utterances among possibly erroneous translation results, the authors use back-translation results to check whether the translation result maintains the semantic meaning of the original sentence. The main difference though is that in the latter paper the method is applied using a very small dataset (less than 3k translated utterances) while we work with around 10 millions.


\section{Method}
\label{sec:methods}

In this paper, we explore bootstrapping of NLU models for a new language by translating training data from an NLU system for a different language. The training data is representative of user requests to voice-controlled assistants; annotations are projected from source to target utterances during MT decoding. 
Since the quality of NLU models trained on MT data depends heavily on the quality of the MT data, we explore different methods for filtering and post-processing. In the following, we describe all approaches in more detail.

\subsection{Filtering}
The goal of the filtering approaches is to choose "good" translations, i.e. we aim to keep primary translations in the training data which are likely to be useful for building NLU models. We explore two approaches for filtering, one based on MT system scores and one based on semantic information.

\subsubsection{Filtering based on semantic information}

 \citet{Misu2012} remove erroneous machine translations in the NLU training data by using back-translations to measure whether the semantic information of a source utterance is retained in the translated utterance. In particular, they apply the following steps:
\begin{enumerate}[itemsep=-1ex]
\item Label the source utterance with an NLU model
\item Translate the source utterance
\item Label the translated utterance by aligning with the result of step 1
\item Translate the translated utterance back into the source language 
\item Label the back-translated utterance with an NLU model
\item Keep the target utterance, if the the recognised intents of steps 1 and 5 are the same
\end{enumerate}
The authors present results with Japanese as the source and English as the target language, suggesting improved spoken language understanding results by filtering translations for the training data with their approach. 
Thus, this approach aims to keep translations for which some semantic information of utterances is retained, potentially avoiding errors in the NLU models trained on these data. We apply this approach in an adapted form, i.e. instead of the additional alignment step (3), we project labels using the MT system, i.e. we make use of the alignment model trained for the MT system. In addition, we extend the approach by 1) determining if the recognised slots are retained in addition to the intent, and 2) making use of the NLU model's confidence, i.e. we remove utterances retaining the intent, if the confidence of the NLU model is very low ($< 0.1$ on a scale from $0-1$). 

\subsubsection{Filtering based on MT scores}
This approach explores the scores returned by the MT system for choosing translations from a training dataset. Since annotating the translations for quality judgement by humans is expensive, we considered to use the translation score as a quality metric that can give us relative quality judgement among a list of translations. In particular, we computed a threshold for each domain based on translation scores. The score we used is the weighted overall translation score as given by Moses MT toolkit and combining the scores of the translation model, the language model, the reordering score and some word penalty. To create a domain-wise threshold, given a translated utterance and its score, we first normalised the score by utterance length. Afterwards, we computed mean and standard deviation per domain. We then selected translations that have a score greater than or equal to the threshold. In this work, we evaluated different thresholds like mean of the translation scores, mean+stdev (standard deviation), mean+(0.5*stdev), and mean+(0.25*stdev).

\subsection{Language-specific post-processing}
Aiming to improve the quality of slot values in the translated data, we explore two strategies for language-specific post-processing. 

\subsubsection{Slot resampling}
If data are translated from another language, slot values related to the countries in the source language might not model those of user requests in the target language. For example, when requesting a weather forecast, American customers would much more frequently ask for an American city than a German one. Thus, an utterance "how is the weather in New York" is likely to be much more frequent in the resulting training data than an utterance "how is the weather in Berlin", and consequently it would appear more frequently in the data after translation to German. This, however, doesn't seem to model language use by German customers well and can hence potentially degrade performance of statistical models trained on these data. 
Aiming to decrease the mismatch in slots values between source language and target language use, we used catalogs to resample slot values for slots where this seemed to be appropriate. In particular, we replaced slot values in the translated data using target language catalog entries corresponding to the slot. For instance, a catalog with German cities can be used to replace "New York" by "Berlin" in the previously mentioned example. For catalogs comprising information, which can be used for weighting catalog entities, we made use of it in that we sample entities according to weights, i.e. the higher the weight, the more often the corresponding entity is sampled. For example, the number of orders can be used to weight albums and population size can be used to weight cities. 

\subsubsection{Keeping some original slot values}

Machine translation systems might incorrectly translate slot values which should not be translated. For example, in an utterance "play we are the champions by queen", the song title "we are the champions" and the band name "queen" should not be translated. While we can apply slot resampling to ingest existing slot values into such utterances, we also explore a different approach. In particular, in this approach we post-process the translated utterances to retain the slot values from the source language utterances for certain slots, such as artists or song titles.  

\section{Experimental setup}
\label{sec:exp}

We ran experiments using US English as the source and German as the target language. Since we are interested in bootstrapping an NLU model for a new language which would first be deployed to beta customers, we evaluate our approach on German beta data. In the following, we first briefly describe the MT and NLU systems and subsequently the datasets.

\subsection{MT and NLU systems}
We used a phrase-based MT system which was built using Moses \cite{Koehn2007} for a similar task, i.e. Question Answering (QA). The MT system is a multi-domain model trained on a mixture of internal and external parallel data sources, which are not from the QA domain; overall the out-of-domain data sources comprised 28,733,606 segments. The  system was fine-tuned using a small manually created parallel corpus for QA, comprising 4,000 segments, and 424,921 in-domain target language segments were used for the target language model. Training data were pre-processed, in particular they were converted into spoken form before building the MT system to better match spoken user utterances of an NLU system. We used an MT system for a similar task rather than an MT system adapted for our data, because we would expect that suitable in-domain data for adaptation might not yet be available for early bootstrapping, i.e. when target language data have not yet been collected. 

For building NLU models, we use Conditional Random Fields \cite{Lafferty2001, CRFSuite} for Named Entity Recognition and a Maximum Entropy classifier \cite{Berger1996} for Intent Classification; we keep the sets of features, hyper-parameters and configuration constant for our experiments. 

\subsection{Datasets}
We translated 10M of training data utterances from a US English NLU system. Overall, the data cover several domains with a large number of different intents and slots/named entities
. We translated the data using the previously described MT system. NLU labels were kept and aligned during the MT decoding to project them from the English source utterances to the corresponding German translations. The final training dataset comprised 9,963,624 utterances. 

For testing, we created a dataset collected from German Beta users; German test data were manually transcribed and annotated with intents and slots/named entities. The resulting test dataset comprised 119,772 utterances. \\
To create a baseline, we used an in-house data collection of 10k utterances, which were manually transcribed and annotated with intents and slots. While in-house data collections are costly and time-consuming, they constitute a reasonable approach for bootstrapping a model from scratch when customer data are not yet available. 

The data amounts are summarised in Table \ref{data}.
\begin{table}[ht]
\centering
\begin{tabular}{l r}
\textbf{Dataset} & \textbf{No. utt.} \\
\hline 
US EN$\rightarrow$DE translated data (train) & 9,963,624\\
DE Beta data (test) &119,772 \\
In-house data collection (train) & 10,000\\
\hline
\end{tabular}\caption{Number of utterances per dataset.}\label{data}
\end{table}

In addition, we created a grammar-based baseline. In particular, we randomly sampled utterances from grammars and created a training dataset from them. For this, we used around 200 grammars written by language experts covering (most) intents and slots supported by the NLU system. However, one of the domains was not covered by grammars, because it supports very diverse features and requests, which are difficult to capture by a grammar.  

We report results by means of a semantic error rate (SemER) which is computed based on the number of insertions, deletions and substitutions for slots and the intent in a recognised utterance compared to a reference utterance, i.e. 

$\text{SemER} = \frac{\text{\# (slot + intent errors)}}{\text{\# (slots + intents in reference)}}$

\section{Results}
\label{sec:res}

First, we compare our approach to the baseline approaches based on grammars and an in-house data collection. For this, we trained NLU models on MT data, on the in-house data collection, on grammar-generated data as well as on MT data together with each baseline dataset. Subsequently, we evaluated the models on the German beta data test set.
Results for model trained on the MT dataset and on the baseline datasets are presented in Table \ref{grammar-vs-mt}.

\begin{table}[ht]
\centering
\begin{tabular}{l c}
\textbf{Training data} & \textbf{SemER (\%)} \\
\hline
Grammar-generated data (baseline) & 55.44\\
In-house collection (baseline) & 23.30\\
\hline 
MT data & 21.38\\
MT data $+$ grammar-generated data & 20.00\\
MT data  $+$ in-house collection & 17.20\\
\hline
\end{tabular}\caption{Comparison for NLU models trained on MT data to our baseline models, i.e. models trained on grammar-generated data and models trained on an in-house data collection of 10k utterances.}\label{grammar-vs-mt}
\end{table}

\begin{table*}
\centering
\begin{tabular}{l c l}
\textbf{Approach(es)} & \textbf{Dataset size}&\textbf{SemER (\%)} \\
\hline
Translated data (baseline) & 9,963,624 & 21.38\\
\hline
Sem. filtering, intents only & 6,694,739 & 20.72 (-3.10)  \\
Sem. filtering, intents $+$ slots & 6,194,498 & 20.60 (-3.64)  \\
Sem. filtering, intents, exclude low confidence & 6,500,127 &20.32 (-4.97)\\
\hline
Filtering based on MT scores, mean & 5,281,331 & 23.62 (+10.48)\\
Filtering based on MT scores, standard deviation  & 8,798,330 &  21.92 (+2.50)\\
Filtering based on MT scores, standard deviation, 0.25  & 6,286,603 & 21.05 (-1.54)\\
Filtering based on MT scores, standard deviation, 0.5  & 7,547,861 & 23.24 (+8.68)\\
\hline
\end{tabular}\caption{Results of the filtering approaches. Relative changes with respect to the baseline are given in parentheses.}\label{filtering}
\end{table*}

As can be seen, the MT approach outperforms the grammar-based one by a large percentage (i.e. around 33\% absolute in SemER), while requiring much less manual effort. In addition, training on both MT and grammar-generated data improves performance over training solely on either one of the datasets; the improvement of the joined approach is particularly large over training solely on grammar-generated data (i.e. around 35 \% absolute in SemER). As noted before, the grammars did not cover one of the domains, yielding errors for its test utterances. To get an estimate of this impact, we removed all utterances from this domain from the test set and recomputed SemER for the grammar-based baseline. While SemER dropped to 34.23, there is still a large difference in performance compared to training on MT data and one domain is not supported at all, even though it is needed by the system.  Compared to the grammar-based baseline, training on the in-house data collection yields a lower SemER of  23.3. Still, training on MT data outperforms this baseline as well, and combining MT data with the in-house data collection improves further over training solely on either one of the datasets (i.e. 17.2 for both vs 21.38 for MT and 23.3 for the in-house collection). Thus, MT data appear to be useful for both bootstrapping an NLU model from scratch and enhancing models trained on grammar-generated data or on an in-house data collection of 10k.

In the following, we evaluate whether our filtering and post-processing approaches can improve the quality of the MT training data further. Table \ref{filtering} presents the results for our filtering approaches. While filtering based on semantic information yields an improvement in SemER over using MT data as they are, filtering based on MT scores only yields a slight improvement in one of the conditions. For filtering based on semantic information, our results confirm in a large-scale industry setting that training on utterances for which the intent is retained after back-translation is useful. In addition, our results show that performance can be improved further by either additionally removing utterances for which the slots are not retained or by removing utterances for which the confidence is very low. Here, removing low confidence utterances yields slightly better results. While we tested with $0.1$ as a threshold, results might be improved further by optimising this threshold, potentially even per domain.

Filtering based on MT scores decreases performance in all considered conditions, except mean+(0.25*stdev), which yields a very slight improvement. However, results were not consistent across domains, i.e., while overall SemER as well as SemER for several domains increased in most cases, it decreased for several domains with relative decreases of up to 48.35\%. Here, manual inspection of the data indicates that this approach is not well-suited for domains comprising very diverse data, since one threshold based on a mean score cannot capture diverse data well. In addition, manual inspection revealed that a rather large percentage of the increase in SemER was due to removing ambiguous utterances, as these typically have a rather low MT score, but are sometimes very frequently used and hence need to be captured by the NLU system. For example, the German utterance "weiter" is frequently used, but can mean both "forward" and "resume" in English, implying also different user intents. Removing only this one utterance from the training data yielded around 2.5k errors, since this utterance is frequent in the test data. However, frequent errors could potentially be fixed manually with little effort by a rule-based approach. Since the approach based on MT scores yields improvements for certain domains, further investigations on the nature of datasets/domains it works well with could be interesting. Since it also yields improved results for some domains compared to the approach based on semantic information, further experiments investigating the combination of both approaches could potentially improve results further. 
 
Table \ref{postprocessing} shows the results of our language-specific post-processing approaches.

\begin{table}
\centering
\begin{tabular}{l l}
\textbf{Approach(es)} & \textbf{SemER (\%)} \\
\hline
Translated data (baseline) & 21.38\\
\hline 
Slot resampling &  21.53 (+0.69) \\
Original slots & 23.82 (+11.40)\\
Original slots $+$ resampling & 20.18 (-5.60)\\
\hline
\end{tabular}\caption{Results of the language-specific post-processing approaches. Relative changes in relation to the baseline are given in parentheses.}\label{postprocessing}
\end{table}

The results show that slot resampling has almost no impact on the error rates. The reason might be that the statistical model uses catalogs also as gazetteers, and hence already includes information on German entities during training. Future work might investigate the effect of slot resampling for models which do not use gazetteers. 
Keeping some original slot values degrades performance from 21.38\% to 23.82\%. One reason for the decrease in performance might be that keeping a few original slot values decreases the frequency of appearance of some German words that still appear in the test set and are requested by users in German. However, it is not consistent that some words or some slots are always in German or English, yielding some mismatches between translated training data and test data. Aiming to counterbalance the increase of English words' frequency, but also consider the original slot values, we combined both approaches. As can be seen in the table, the combined approach yields an improvement, i.e. SemER is 20.18\% compared to  21.38\% for training on MT data as they are. With 20.18\%, this approach is also performing better than our best-performing filtering approach which yields 20.32\% in SemER. One interesting question for future work will be to explore if combining filtering and language-specific post-processing approaches will improve results further.\\
Overall, compared to the grammar-based baseline, the best-performing post-processing approach yields a large improvement in SemER (20.18\%  vs 55.44\%) and also yields results very similar to the NLU model trained on both the initial MT data and the grammar-generated data (20.18\% vs 20.0\%). However, our proposed post-processing approach can be applied automatically and quickly, while grammar writing is very costly and time-consuming. 

\section{Conclusion}
\label{sec:concl}

Aiming to reduce time and costs needed to bootstrap an NLU model for a new language, in this paper we made use of MT data to build NLU models. In addition, we compared different techniques to filter and post-process the MT data, aiming to improve NLU performance further. These methods were evaluated in large-scale experiments for a voice-controlled assistant to bootstrap a German system using English data. The results when using MT data showed a large improvement in performance compared to a grammar-based baseline and outperformed a baseline using an in-house data collection. The applied filtering and post-processing techniques improved results further over using MT data as they are.\\
In future work, we plan to apply our approach to further languages and explore bootstrapping new domains for an existing NLU system.

\section*{Acknowledgments}
We thank Steve Sloto, Greg Hanneman, Donna Gates and Patrick Porte for building the MT system and Fabian Triefenbach and Daniel Marcu for valuable comments on this paper.
\bibliography{naaclhlt2018}
\bibliographystyle{acl_natbib}

\end{document}